\documentclass[runningheads]{llncs}
\usepackage{graphicx}
\usepackage{amsmath,amssymb} 
\usepackage{color}

\usepackage{comment}
\usepackage{float}
\usepackage{times}
\usepackage{epsfig}
\usepackage{graphicx}
\usepackage{amsmath}
\usepackage{amssymb}
\usepackage{footnote}
\usepackage{mathtools}
\usepackage{caption}
\usepackage{booktabs} 
\usepackage[utf8]{inputenc}
\usepackage[english]{babel}
\usepackage{tablefootnote}
\usepackage{cite}
\usepackage[table]{xcolor}
\usepackage{multirow}

\newcommand{\etal}{\textit{et al}.}

\begin{document}
\pagestyle{headings}
\mainmatter

\title{Question Type Guided Attention in Visual Question Answering} 

\titlerunning{Question Type Guided Attention in Visual Question Answering}

\authorrunning{Y. Shi and T. Furlanello and S. Zha and A. Anandkumar}

\author{Yang Shi\inst{1}  \thanks{Work partially done while the author was working at Amazon AI}
\and Tommaso Furlanello \inst{2}
 \and Sheng Zha\inst{3} \and Animashree Anandkumar \inst{3}\inst{4}}

\institute{University of California, Irvine \\
\email{shiy4@uci.edu}
\and
University of Southern California \\
\email{furlanel@usc.edu}
\and Amazon AI \\
\email{\{zhasheng\},\{anima\}@amazon.com}
\and California Institute of Technology}


\maketitle

\begin{abstract}
Visual Question Answering (VQA) requires  integration of feature maps with drastically different structures. Image descriptors have structures at multiple spatial scales, while lexical inputs inherently follow a temporal sequence and naturally cluster into semantically different question types. A lot of previous works use complex models to extract feature representations but neglect to use high-level information summary such as question types in learning.
In this work, we propose Question Type-guided  Attention (QTA). It utilizes the information of question type to dynamically balance between bottom-up and top-down visual features, respectively extracted from ResNet and Faster R-CNN networks.
We experiment with multiple VQA architectures with extensive input ablation studies over the TDIUC dataset and show that QTA systematically improves the performance by more than 5\% across multiple question type categories such as ``Activity Recognition'', ``Utility'' and ``Counting'' on TDIUC dataset compared to the state-of-art. By adding QTA on the state-of-art model MCB, we achieve 3\% improvement in overall accuracy. 
Finally, we propose a multi-task extension to predict question types which generalizes QTA to applications that lack question type, with a minimal performance loss.
\end{abstract}

\section{Introduction}
The relative maturity and flexibility of deep learning allow to build upon the success of computer vision~\cite{alexnet}
and natural language~\cite{lstm, w2v}
to face new complex and multimodal tasks. Visual Question Answering(VQA)~\cite{paper:VQA} focus on providing a natural language answer given any image and any free-form natural language question. 
To achieve this goal, information from multiple modalities must be integrated. Visual and lexical inputs are first processed using specialized encoding modules and then integrated through differentiable operators. 
Image features are usually extracted by convolution neural networks~\cite{DBLP:journals/corr/DonahueJVHZTD13}, while recurrent neural networks~\cite{DBLP:journals/corr/SutskeverVL14, lstm} are used to extract question features.
Additionally, attention mechanism~\cite{attention,yang,imagecaption} forces the system to \emph{look at} informative regions in both text and vision. Attention weight is calculated from the correlation between language and vision features and then is multiplied to the original feature.

Previous works explore new features to represent vision and language. 
Pre-trained ResNet~\cite{resnet} and VGG~\cite{vgg} are commonly used in VQA vision feature extraction. The authors in~\cite{winnercvpr} show that post-processing CNN with region-specific image features~\cite{bottomupattention} such as Faster R-CNN~\cite{fasterrcnn} can lead to an improvement of VQA performance. Along with generating language feature from either sentence-level or word-level using LSTM~\cite{lstm} or word embedding, Lu \etal~\cite{coatt} propose to model the question from word-level, phrase-level, and entire question-level in a hierarchical fashion.
 
Through extensive experimentation and ablation studies, we notice that the role of “raw” visual features from ResNet and processed region-specific features from Faster R-CNN is complementary and leads to improvement over different subsets of question types. However, we also notice that trivial  information in VQA dataset: question/answer type is omitted in training. Generally, each sample in any VQA dataset contains one image file, one natural language question/answer and sometimes answer type. 
A lot of work use the answer type to analyze accuracy per type in result~\cite{paper:VQA} but neglect to use it during learning. TDIUC~\cite{TDIUC} is a recently released dataset that contains question type for each sample. Compared to answer type, question type has less variety and is easier to interpret when we only have the question. 

\begin{figure}[H]
\centering
\includegraphics[width=10cm,height = 4cm]{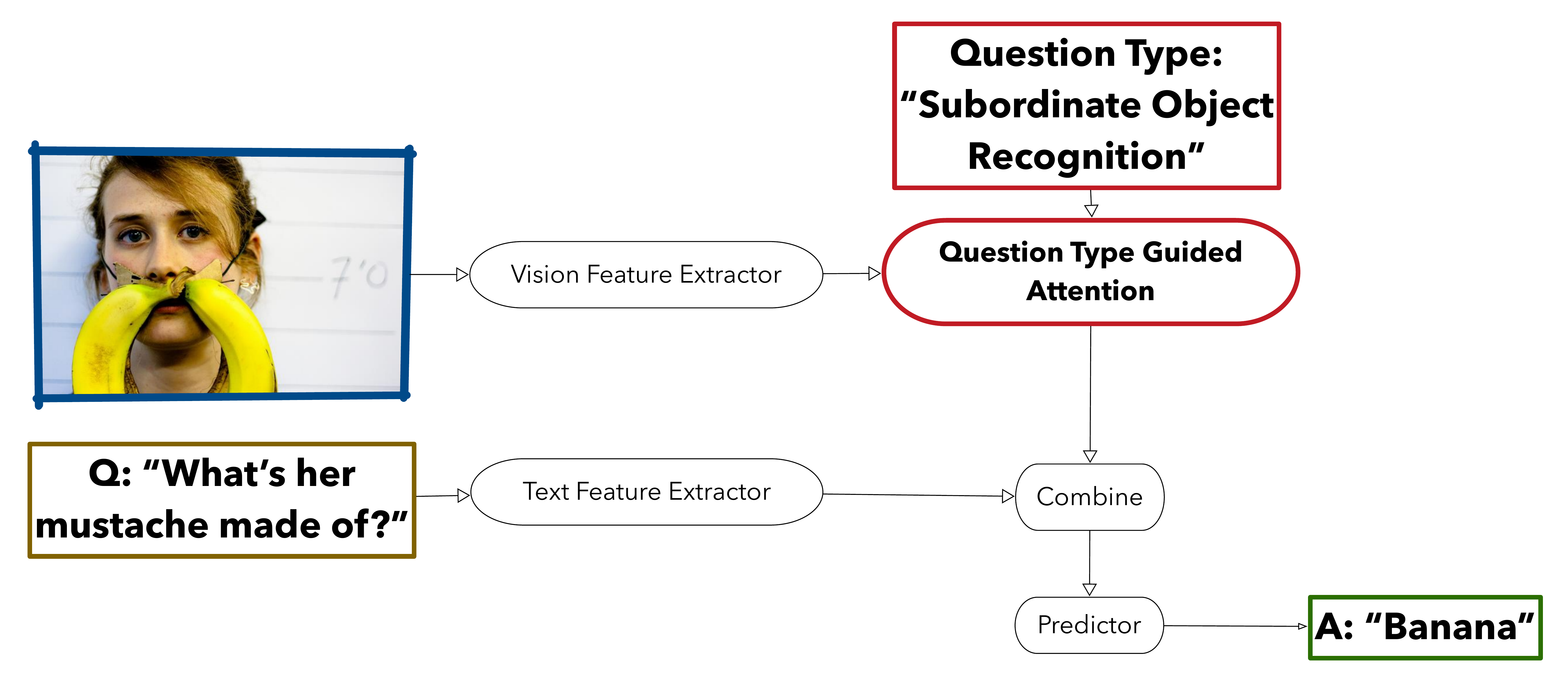}
\caption{General VQA network with QTA}
\label{fig:vqa-basic}
\end{figure}

The focus of this work is the development of an attention mechanism that exploits high-level semantic information on the question type to guide the visual encoding process. This procedure introduces information leakage between modalities before the classical integration phase that improves the performance on VQA task. Specifically, We introduce a novel VQA architecture \textbf{Question Type-guided Attention}(QTA) that dynamically gates the contribution of ResNet and Faster R-CNN features based on the question type. Our results with QTA allow us to integrate the information from multiple visual sources and obtain gains across all question types. A general VQA network with our QTA is shown in Figure~\ref{fig:vqa-basic}. 

The contributions of this paper are:(1) We propose question type-guided attention to balance between bottom-up and top-down visual features, which are respectively extracted from ResNet and Faster R-CNN networks. Our results show that QTA systematically improves the performance by more than 5\% across multiple question type categories such as ``Activity Recognition'', ``Utility'' and ``Counting'' on TDIUC dataset. By adding QTA to the state-of-art model MCB, we achieve 3\% improvement in overall accuracy. (2)
We propose a multi-task extension that is trained to predict question types from the lexical inputs during training time that do not require ground truth labels during inference. We get more than 95\% accuracy for the question type prediction while keeping the VQA task accuracy almost same as before. (3) Our analysis reveals some problems in the TDIUC VQA dataset. Though the ``Absurd'' question is intended to help reduce bias, it contains too many similar questions, specifically, questions regarding color. This will mislead the machine to predict wrong question types. Our QTA model gets 17\% improvement on simple accuracy compared to the baseline in~\cite{TDIUC} when we exclude absurd questions in training.

\section{Related Works}
VQA task is first proposed in~\cite{paper:VQA}.
It focuses on providing a natural language answer given any image and any free-form natural language question. Collecting data and solving the task are equally challenging as they require the understanding of the joint relation between image and language without any bias.
\label{sec:related}

\textbf{Datasets}
VQA dataset v1 is first released by Antol \etal~\cite{paper:VQA}. The dataset consists of two subsets: real images and abstract scenes. However, the inherent structure of our world is biased and it results in a biased dataset. In another word, a specific question tends to have the same answer regardless of the image. For example, when people ask about the color of the sky, the answer is most likely blue or black. It is unusual to see the answer be yellow. This is the bottleneck when we give a yellow color sky and ask the machine to answer it. Goyal \etal~\cite{balanced_vqa_v2} release VQA dataset v2. This dataset pairs the same question with similar images that lead to different answers to reduce the sample bias. 
Agrawal \etal ~\cite{dontassume} also noticed that every question type has different prior distributions of answers. Based on that they propose GVQA and new splits of the VQA v1/v2. In the new split, the distribution of answers per question type is different in the test data compared to the training data.
Zhang \etal~\cite{balanced_binary_vqa,zhang-thesis} also propose a method to reduce bias in abstract scenes dataset at question level. By extracting representative word tuples from questions, they can identify and control the balance for each question. 
Vizwiz~\cite{vizwiz} is another recently released dataset that uses pictures taken by blind people. Some pictures are of poor quality, and the questions are spoken. These data collection methods help reduce bias in the dataset.

Johnson \etal~\cite{CLVR} introduce Compositional Language and Elementary Visual Reasoning (CLEVR) diagnostic dataset that focuses on reasoning. Strub \etal~\cite{guesswhat} propose a two-player guessing game: guess a target in a given image with a sequence of questions and answers. This requires both visual question reasoning and spatial reasoning.

The Task Driven Image Understanding Challenge dataset(TDIUC)~\cite{TDIUC} contains a total of over 1.6 million questions in 12 different types. It contains images and annotations from MSCOCO~\cite{mscoco} and Visual genome~\cite{visualgenome}. The key difference between TDIUC and the previous VQA v1/v2 dataset is the categorization of questions: Each question belongs to one of the 12 categories. This allows a task-oriented evaluation such as per question-type accuracies. They also include an ``Absurd" question category in which questions are irrelevant to the image contents to help balance the dataset. 

\textbf{Feature Selection}
VQA requires solving several tasks at once involving both visual and textual input: visual perception, question understanding, and reasoning. Usually, features are extracted respectively with convolutional neural networks~\cite{DBLP:journals/corr/DonahueJVHZTD13} from the image, and with recurrent neural networks~\cite{DBLP:journals/corr/SutskeverVL14, lstm} from the text. 

Pre-trained ResNet and VGG are commonly used in VQA vision feature extraction. The authors in~\cite{winnercvpr} show that post-processing CNN with region-specific image features~\cite{bottomupattention} can lead to an improvement of VQA performance. Specifically, they use pre-trained Faster R-CNN model to extract image features for VQA task. They won the VQA challenge 2017. 

On the language side, pre-trained word embeddings such as Word2Vec~\cite{w2v} are used for text feature extraction. There is a discussion about the sufficiency of language input for VQA task. Agrawal \etal~\cite{ABVQAM} have shown that state-of-art VQA models converge to the same answer even if only given half of the question compared to if given the whole sentence.

\textbf{Generic Methods}
Information of both modalities are used jointly through means of combination, such as concatenation, product or sum. In~\cite{paper:VQA}, authors propose a baseline that combines LSTM embedding of the question and CNN embedding of the image via a point-wise multiplication followed by a multi-layer perceptron classifier.

\textbf{Pooling Methods}
Pooling methods are widely used in visual tasks to combine information for various streams into one final feature representation. Common pooling methods such as average pooling and max pooling bring the property of translation invariance and robustness to elastic distortions at the cost of spatial locality. Bilinear pooling can preserve spatial information, which is performed with the outer product between two feature maps. However, this operation entails high output dimension($O(MN)$ for feature maps of dimension $M$ and $N$). This exponential growth with respect to the number of feature maps renders it too costly to be applied to huge real image datasets. There have been several proposals for new pooling techniques to address this problem:
\begin{itemize}
\item
Count sketch~\cite{countsketch} is applied as a feature hashing operator to avoid dimension expanding in bilinear pooling. 
Given a vector $a \in \mathcal{R}^n$, 
random hash function $f \in \mathcal{R}^n$: $[n] \to [b]$ and binary variable $s \in \mathcal{R}^n$: $[n] \to \pm 1$, the \textbf{count sketch}~\cite{countsketch} operator $cs(a,h,s) \in \mathcal{R}^b$ is:
\vspace*{-5pt}
\begin{align}
cs(a,f,s)[j] = \sum_{f[i] = j}s[i]a[i], \quad j \in {1,\cdots,b}
\end{align}
Gao \etal~\cite{CBP} use convolution layers from two different neural networks as the local descriptor extractors of the image and combine them using count sketch. ``$\alpha$-pooling"~\cite{alphapooling} allows the network to learn the pooling strategy: a continuous transition between linear and polynomial pooling. They show that higher $\alpha$ gives larger gain for fine-grained image recognition tasks. However, as $\alpha$ goes up, the computation complexity increases in polynomial order. 
\item
Fukui \etal~\cite{MCB} use count sketch as a pooling method in VQA tasks and obtains the best results on VQA dataset v1 in VQA challenge 2016. They compute count sketch approximation of the visual and textual representation at each spatial location. 
Given text feature $v \in \mathcal{R}^{L}$ and image features $I \in \mathcal{R}^{C \times H \times W}$, Fukui \etal~\cite{MCB} propose \textbf{MCB} as:
\begin{align}
MCB&(I[:,h,w] \otimes v)[t_1, h,w] \nonumber \\
& = (cs(I[:,h,w],f,s) \star cs(v,f,s))[t_1, h,w] \nonumber \\
& = IFFT1(FFT1(cs(I[:,h,w],f,s))[t_1, h,w] \circ FFT1(cs(v,f,s))[t_1] )\label{eqn:3} \nonumber \\ 
&h \in \{1,\cdots H\}, w \in \{1, \cdots W\}, t_1 \in \{1,\cdots, b\} 
\end{align}
$\otimes$ denotes outer product. $\circ$ denotes element-wise product. $\star$ denotes convolution operator. This procedure preserves spatial information in the image feature.
\end{itemize}
\textbf{Attention}
Focusing on the objects in the image that are related to the question is the key to understand the correlation between the image and the question. Attention mechanism is used to address this problem. There are soft attention and hard attention~\cite{imagecaption} based on whether the attention term/loss function is differentiable or not. Yang \etal~\cite{yang} and Xu \etal~\cite{attention} propose word guided spatial attention specifically for VQA task. Attention weight at each spatial location is calculated by the correlation between the embedded question feature and the embedded visual features. The attended pixels are at the maximum correlations.  Wang \etal~\cite{pos-attention} explore mechanisms of triplet attention that interact between the image, question and candidate answers based on image-question pairs.

\section{Question Type Guided Visual Attention}
\label{sec:qta}
Question type is very important in predicting the answer regardless whether we have the corresponding image or not. For example, questions starting with ``how many" will mostly lead to numerical answers. Agrawal \etal~\cite{ABVQAM} have shown that state-of-art VQA models converge to the same answer even if only given half of the question compared to if given the whole sentence. Besides that, inspired by~\cite{winnercvpr}, we are curious about combining bottom-up and top-down visual features in VQA task. To get a deep understanding of visual feature preference for different questions, we try to find an attention mechanism between these two. Since question type is representing the question, we propose Question Type-guided Attention(QTA).

Given several independent image features $F_1, F_2, \cdots F_k$, such as features from ResNet, VGG or Faster R-CNN, we concatenate them as one image feature: $F = [F_1,F_2, \cdots F_k] \in \mathcal{R}^{M}$. Assume there are $N$ different question types, QTA is defined as $F \circ WQ$, where $Q \in \mathcal{R}^{N}$ is the one-hot encoding of the question type, and $W \in \mathcal{R}^{M \times N}$ is the hidden weight. We can learn the weight by back propagation through the network. In other words, we learn a question type embedding and use it as attention weight.

\begin{figure}[]
\centering
\begin{minipage}{.5\textwidth}
  \centering
\includegraphics[width=6cm,height = 4cm]{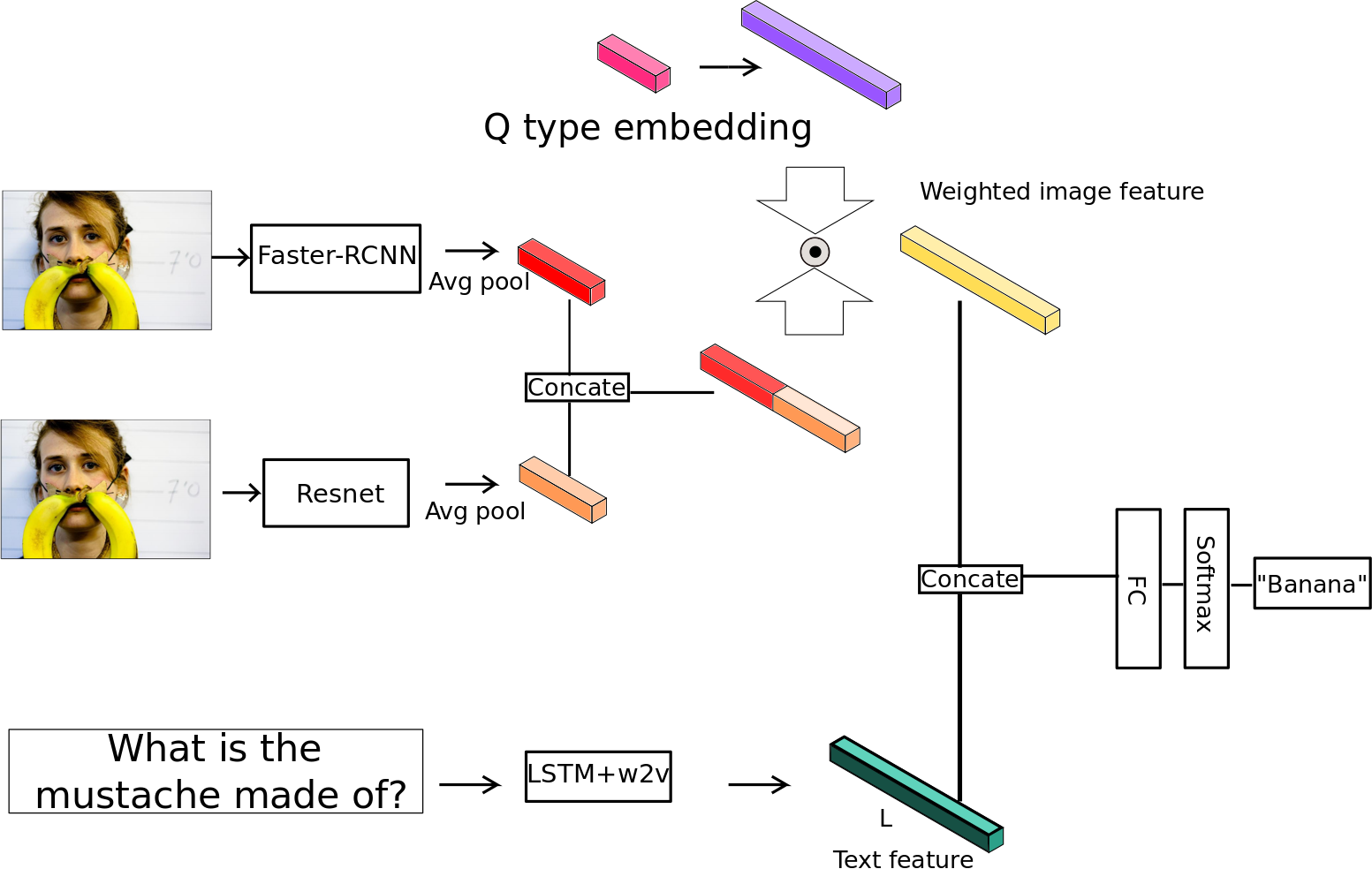}
\captionof{figure}{Concatenation model with QTA structure for VQA task(CATL-QTA$^W$ in Section~\ref{sec:experiment})}
\label{fig:qtype}
\end{minipage}%
\hspace*{5pt}
\begin{minipage}{.4\textwidth}
  \centering
  \includegraphics[width=5.7cm,height = 4cm]{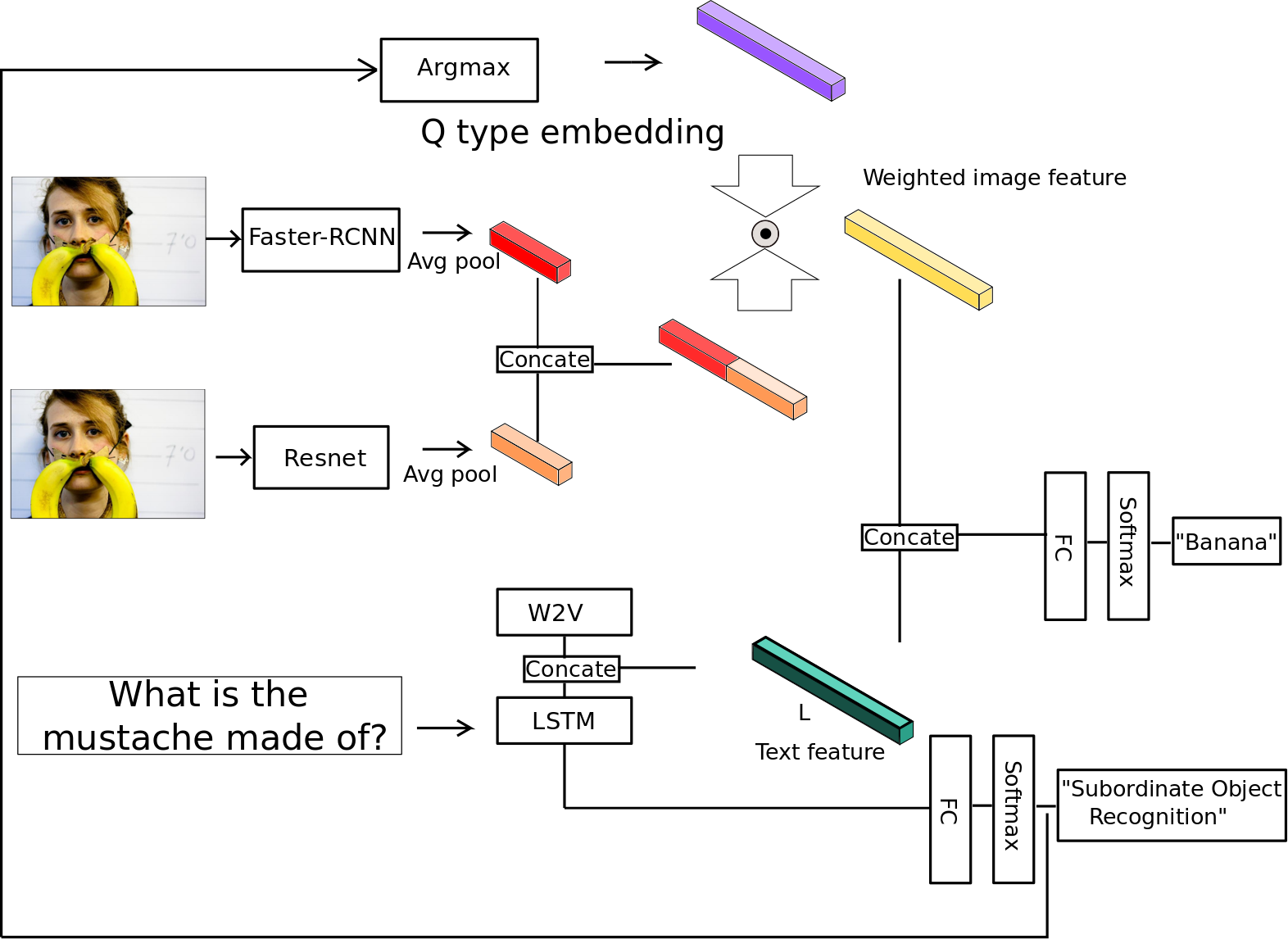}
\captionof{figure}{Concatenation model with QTA structure for multi-task(CATL-QTA-M$^W$ in Section~\ref{sec:experiment})}
\label{fig:multitask-qtype}
\end{minipage}
\end{figure}

\begin{figure}[]
\centering
\includegraphics[width=12cm,height = 5cm]{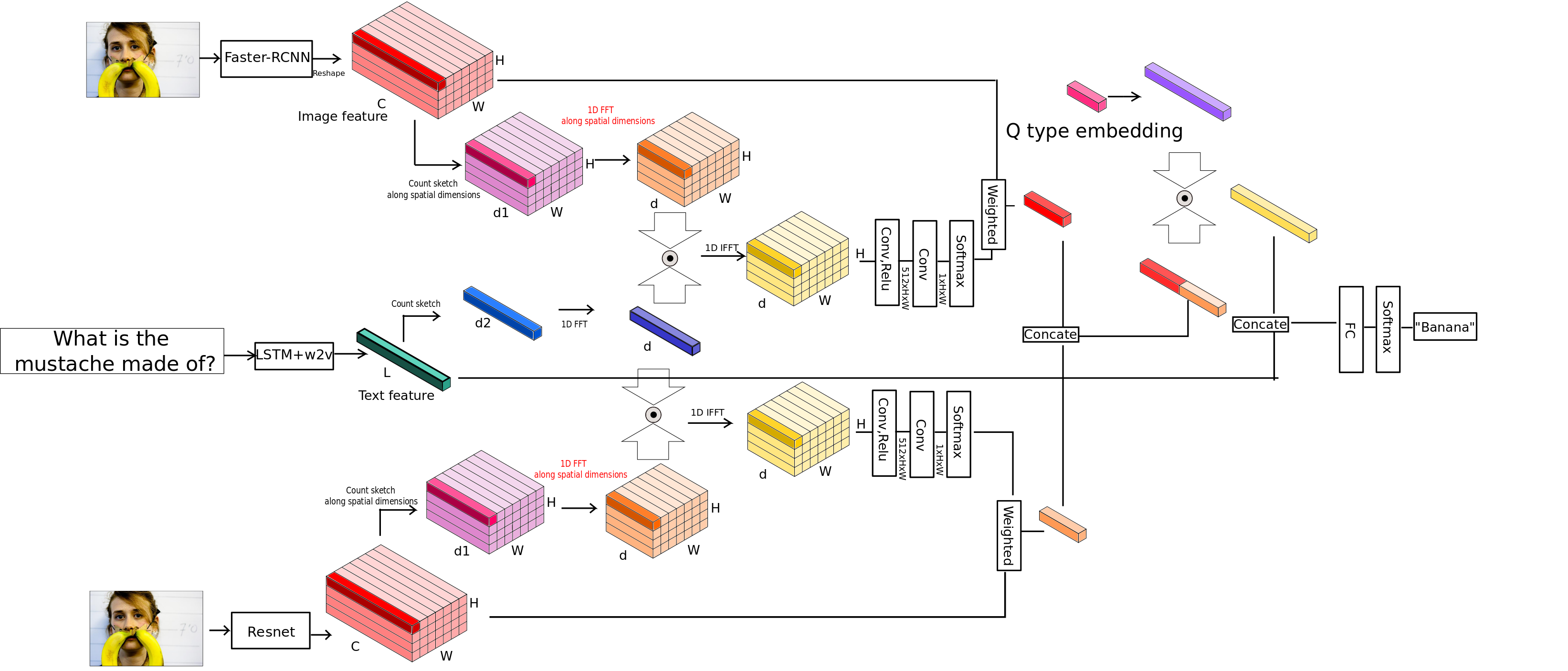}
\caption{MCB model with QTA structure(MCB-QTA in Section~\ref{sec:experiment})}
\label{fig:mcbqtype}
\end{figure}

QTA can be used in both generic and complex pooling models. In Figure~\ref{fig:qtype}, we show a simple concatenation model with question type as input. We describe it in detail in Section~\ref{sec:experiment}. To fully exploit image features in different channels and preserve spatial information, we also propose MCB with question type-guided image attention in Figure~\ref{fig:mcbqtype}. 

One obvious limitation of QTA is that it requires question type label. In the real world scenario, the question type for each question may not be available. In this case, it is still possible to predict the question type from the text, and use it as input to the QTA network. Thus, we propose a multi-task model that focuses on VQA task along with the prediction of the question type in Figure~\ref{fig:multitask-qtype}. This model operates in the setting where true question type is available only at training time. In Section~\ref{sec:result}, we also show through experiment that it is a relatively easy task to predict the question type from question text, and thus making our method generalizable to those VQA settings that lack question type.

\section{Experiments}
\label{sec:experiment}
\vspace*{-5pt}
In this section, we describe the dataset in Section~\ref{sec:expdataset}, evaluation metrics in Section~\ref{sec:expeval}, model features in Section~\ref{sec:expfeat}, and model structures are explained in Section~\ref{sec:4.4}.

\vspace*{-5pt}
\subsection{Dataset}
\vspace*{-5pt}
\label{sec:expdataset}
Our experiments are conducted on the Task Driven Image Understanding Challenge dataset(TDIUC)~\cite{TDIUC}, which contains over 1.6 million questions in 12 different types. This dataset includes VQA v1 and Visual Genome, with a total of 122429 training images and 57565 test images. The annotation sources are MSCOCO (VQA v1), Visual genome annotations, and manual annotations. TDIUC introduces absurd questions that force an algorithm to determine if a question is valid for a given image. There are 1115299 total training questions and 538543 total test questions. The total number of samples is 3 times larger than that in VQA v1 dataset.
\vspace*{-5pt}
\subsection{Evaluation Metrics}
\vspace*{-5pt}
\label{sec:expeval}
There are total 12 different question types in TDIUC dataset as we mentioned in Section~\ref{sec:related}. We calculate the simple accuracy for each type separately and also report the arithmetic and harmonic means across all per question-type(MPT) accuracies. 
\vspace*{-5pt}
\subsection{Feature Representation}
\vspace*{-5pt}
\label{sec:expfeat}
\textbf{Image feature}
We use the output of ``pool'' of a  152-layer ResNet as an image feature baseline. The output dimension is $2048 \times 14 \times 14$.  
Faster R-CNN~\cite{fasterrcnn} focuses on object detection and classification. Teney \etal~\cite{winnercvpr} use it to extract object-oriented features for VQA dataset and show better performance compared to the ones using ResNet feature. We fix the number of detected objects to be 36 and extract the image features based on their pre-trained Faster R-CNN model. As a result, the extracted image feature is a $36 \times 2048$ matrix. 
To fit in MCB model, which requires spatial representation, we reshape it into a $6 \times 6 \times 2048$ tensor.

\textbf{Text feature}
We use common word embedding library: 300-dim Word2Vec~\cite{w2v} as a pre-trained text feature: we sum over the word embeddings for all words in the sentence. A two-layer LSTM is used as an end-to-end text feature extractor. We also use the encoder of google neural machine translation(NMT) system~\cite{nmt} as a pre-trained text feature and compare it with Word2Vec. The pre-trained NMT model is trained on UN parallel corpus 1.0 in MXnet~\cite{mxnet}. Its BLEU score is 34. The output dimension of the encoder is $1024$.

\vspace*{-5pt}
\subsection{Models}
\vspace*{-5pt}
\label{sec:4.4}
\subsubsection{Baseline models}
Baseline models are based on a one-layer MLP: A fully connected network classifier with one hidden layer with ReLu non-linearity, followed by a softmax layer. The input is a concatenation of image and text feature. There are 8192 units in the hidden state. 

\begin{table}[]
\caption{Baseline models}
\label{Table:bl}
\begin{center}
\scalebox{0.95}{
\begin{tabular}{c |c |c | c }
\toprule
  \textbf{Name} & \textbf{Image feature} & \textbf{Text feature} & \textbf{Modal} \\
\midrule
\textbf{CAT1} & ResNet/Faster R-CNN vector feature &Skipthought/NMT/Word2Vec pre-trined feature&MLP \\
\hline
\textbf{CAT1L} & ResNet/Faster R-CNN vector feature &End-to-end 2-layer LSTM's last hidden state&MLP \\
\hline
\multirow{2}{*}{\textbf{CATL}} & {Concatenation of ResNet}& \multirow{2}{*}{End-to-end 2-layer LSTM's last hidden state}& \multirow{2}{*}{MLP} \\
&and Faster R-CNN vector features&&\\
\hline
\multirow{2}{*}{\textbf{CAT2}} & {Concatenation of ResNet}& \multirow{2}{*}{Skipthought/NMT/Word2Vec pre-trined feature}& \multirow{2}{*}{MLP} \\
&and Faster R-CNN vector features&&\\            
\bottomrule
\end{tabular}
}
\end{center}
\end{table}

To compare different image and text feature, we have \textbf{CAT1}, \textbf{CAT1L} and \textbf{CATL}. To check the complementarity of different features between ResNet and Faster R-CNN and show how they perform differently across question types, we set up baseline \textbf{CAT2}. In LSTM, the hidden state length is 1024. The word embedding dimension is 300. Detailed definitions are in Table~\ref{Table:bl}.

To further exam and explain our QTA proposal, we use more sophisticate feature integration operators as a strong baseline to compare with.
\textbf{MCB-A}, as we mentioned in Section~\ref{sec:related}, is proposed in ~\cite{MCB}. \textbf{RAU}~\cite{rau} is a framework that combines the embedding, attention and predicts operation together inside a recurrent network. We reference results of these two models from~\cite{TDIUC}.

\subsubsection{QTA models}
From the baseline analysis, we realize that ResNet and Faster R-CNN features are complementary to each other. Using question type as guidance for image feature selection is the key to make image feature stronger. Therefore, we propose QTA networks in MLP model(\textbf{CATL-QTA}) and MCB model(\textbf{MCB-QTA}). The out dimension of the count sketch in the MCB is 8000. The structures are in Figure~\ref{fig:qtype}, \ref{fig:mcbqtype}. The descriptions are in Table~\ref{Table:qta}.

To check whether the model benefits from the QTA mechanism or from added question type information itself, we design a network that only uses question type embedding without attention. \textbf{CAT-QT} and \textbf{CATL-QT} are the two  proposed network using Word2Vec and LSTM lexical feature. 

As mentions in Section~\ref{sec:qta}, we propose a multi-task network for QTA in case we don't have question type label at inference. \textbf{CATL-QTA-M} is a multi-task model based on CATL-QTA. The output of LSTM is connected to a one-layer MLP to predict question type for the input question. The prediction result is then fed into QTA part through argmax. The Multi-task MLP is in Figure~\ref{fig:multitask-qtype}. 

\begin{table}[]
\caption{QTA models}
\label{Table:qta}
\begin{center}
\scalebox{0.75}{
\begin{tabular}{c |c |c | c }
\toprule
  \textbf{Name} & \textbf{Image feature} & \textbf{Text feature} & \textbf{Modal} \\
\midrule
\multirow{2}{*}{\textbf{CATL-QTA}} & {QTA weighted pre-trained vector features}& \multirow{2}{*}{End-to-end 2-layer LSTM's last hidden state}& \multirow{2}{*}{MLP} \\
&from ResNet and Faster R-CNN&&\\
\hline
\multirow{2}{*}{\textbf{MCB-QTA}} & {QTA weighted pre-trained spatial features}& \multirow{2}{*}{End-to-end 2-layer LSTM's last hidden state}& \multirow{2}{*}{ MCB} \\
&from ResNet and Faster R-CNN&&\\
\hline  
\hline
\multirow{2}{*}{\textbf{CAT-QT}} & {Concatenation of ResNet}&{Concatenation of Word2Vec pre-trined feature}& \multirow{2}{*}{MLP} \\
&and Faster R-CNN vector features& and a 1024-dim question type embedding&\\
\hline            
\multirow{2}{*}{\textbf{CATL-QT}} & {Concatenation of ResNet}& {Concatenation of end-to-end 2-layer LSTM's last }& \multirow{2}{*}{MLP} \\
& hidden state and Faster R-CNN vector features& and a 1024-dim question type embedding&\\
\hline 
\hline
\multirow{2}{*}{\textbf{CATL-QTA-M}} & {QTA weighted pre-trained spatial features}& \multirow{2}{*}{End-to-end 2-layer LSTM's last hidden state}& \multirow{2}{*}{Multi-task MLP} \\
&from ResNet and Faster R-CNN&&\\
\hline            
\bottomrule
\end{tabular}
}
\end{center}
\end{table}

\section{Results and Analysis}
\vspace*{-5pt}
\label{sec:result}
We first focus in Sections \ref{subsec:image} and \ref{subsec:text} on results concerning the complementarity of different features across question category types.
For the visual domain, we explore the use of Faster R-CNN and ResNet features, while for the lexical domain we use NMT, LSTM and pre-trained Word2Vec features.
We then analyze the effect of question type both as input and with QTA in VQA tasks in Section~\ref{subsec:concat}. 
Finally, in the remaining subsections, we extend the basic concatenation QTA model to MCB style pooling; introduce question type as both input and output during training such that the network can produce predicted question types during inference; and study more in depth the effect of the question category ``Absurd'' on the overall model performance across categories.

\subsection{Faster R-CNN and ResNet Features}
\label{subsec:image}

Table~\ref{Table:0} reports our extensive ablation analysis of  simple concatenation models using multiple visual and lexical feature sources.
From the results in the second and third columns, we see that overall the model with Faster R-CNN features outperform the one using ResNet features when using NMT features. 
We show in column 4 that the features sources are complementary, and their combination is better across most categories (in bold) with respect to the single source models in columns 2 and 3.
In columns 5,6; 7,8 and 9,10 we replicate the same comparison between ResNet and R-CNN features using more sophisticate models to embed the lexical information. 
We reach more than 10 \% accuracy increase, from 69.53 \% to 80.16 \% using a simple concatenation model with an accurate selection of the feature type.
\begin{table}
\caption{Benchmark results of concatenation models on TDIUC dataset using different image features and pre-trained language feature. 1:  Use ResNet feature and SkipGram feature 2: Use ResNet feature and NMT feature 3: Use Faster R-CNN feature and NMT feature 4: Use ResNet feature and end-to-end LSTM feature 5: Use Faster R-CNN feature and end-to-end LSTM feature. N denotes that additional NMT embedding is concatenated to LSTM output. W denotes that additional Word2Vec embedding is concatenated to LSTM output(Following tables also use the same notation)}
\label{Table:0}
\begin{center}
\scalebox{0.73}{
\begin{tabular}{c|c ||c c| c || c c || c c||c c}
\toprule
Columns & 1 & 2& 3&4 &5 &6&7&8&9&10\\
  Accuracy(\%) & \textbf{CAT1}$^{1}$~\cite{TDIUC}   & \textbf{CAT1}$^2$& \textbf{CAT1}$^3$   & \textbf{CAT2}  & \textbf{CAT1L}$^4$ & \textbf{CAT1L}$^5$ &  \textbf{CAT1L}$^{4N}$&\textbf{CAT1L}$^{5N}$&  \textbf{CAT1L}$^{4W}$&\textbf{CAT1L}$^{5W}$ \\
\midrule
Scene Recognition & 72.19 & 68.51&68.81 &\textbf{69.06 } & 91.62 &\textbf{92.27} &91.16&\textbf{92.33}&91.57&\textbf{92.45}\\
Sport Recognition &  85.16 &89.67& 92.36&\textbf{93.15 } &  90.94 & \textbf{93.84}&89.62&\textbf{93.52}&90.77 &\textbf{94.05}\\
Color Attributes &  43.69 &32.90& 34.35 &\textbf{34.99}&  45.62 &\textbf{49.43}&44.07&\textbf{47.78}&47.33 &\textbf{49.47}\\
Other Attributes &  42.89&38.05& \textbf{39.76}&39.67 & 40.89 &\textbf{43.49} &39.60&\textbf{42.35}& 41.92&\textbf{45.19}\\
Activity Recognition & 24.16 &39.34& 45.75 &\textbf{46.87} & 42.95&\textbf{49.25}&40.12&\textbf{44.11}&42.13 &\textbf{49.25}\\
Positional Reasoning &  25.15 &25.63& 27.16 &\textbf{28.02 }& 26.22  & \textbf{29.35}&24.17&\textbf{27.50}&25.72& \textbf{28.59}\\
Sub. Object Recognition &  80.92 &83.94&85.67 &\textbf{86.78} &82.20& \textbf{85.06} &81.85&\textbf{84.47}& 82.52&\textbf{85.05}\\
Absurd &  96.96& 94.98&  94.77& \textbf{95.82}& \textbf{90.87}&87.10 &\textbf{95.38}&93.28&\textbf{93.59} &91.95\\
Utility and Affordances &  24.56 &25.93& \textbf{27.78} &27.16& 15.43 &\textbf{25.93}&\textbf{25.31}&18.52&16.05 &\textbf{17.28}\\
Object Presence &  69.43 &77.21&77.90 & \textbf{78.29}&89.40&\textbf{91.14} &90.13&\textbf{91.95} &91.08&\textbf{91.81}\\
Counting &  44.82&48.46& 52.18 & \textbf{52.57}& 45.95 &\textbf{50.27}&44.26 &\textbf{49.24}&44.93&\textbf{51.30}\\
Sentiment Understanding &53.00 &43.45& 46.49& \textbf{47.28}&46.49& \textbf{48.72}&41.85&\textbf{42.81}&44.89&\textbf{46.01}\\
\hline
Overall (Arithmetic MPT) & 55.25&55.67&57.57&58.31&59.05 & 62.15&58.96&60.66&59.38 &\textbf{61.80}\\
Overall (Harmonic MPT) & 44.13&45.37&47.99&48.44&44.09&\textbf{51.66}&46.84& 46.84& 44.42&47.70\\
\hline
Overall Accuracy &69.53&71.41&72.44& 73.05 & 77.55&78.66&78.35&79.94&78.94&\textbf{80.16}\\
\bottomrule
\end{tabular}
}
\end{center}
\end{table}

\vspace*{-5pt}
\subsection{Pre-trained and Jointly-trained Text Feature Extractors}
\vspace*{-5pt}
\label{subsec:text}
The first four columns in Table~\ref{Table:0} show the results of models with text features from NMT. To fully explore the text feature extractor in VQA system, we substitute the NMT pre-trained language feature extractor with a jointly-trained two layer LSTM model. The improved performance of jointly-training text feature extractor can be appreciated by comparing the results of the 4 left-most and right most columns in Table~\ref{Table:0}. For example, comparing second column and fifth column in Table~\ref{Table:0}, we get 6\% improvement using LSTM while keeping image feature and network same.

We obtain the best model by concatenating the output of the LSTM and the pre-trained NMT/Word2Vec feature, as shown in Table~\ref{Table:0}. It gives us $10\%$ improvement for ``Utility and Affordances'' when we look at the fifth and seventh column. We find the use of Word2Vec is better than NMT feature in last four columns in Table~\ref{Table:0}. We think the better performance of Word2Vec with respect to the NMT encoder, might be due to the more similar structure of single sentence samples of  Word2Vec  training set with those from classical VQA dataset with respect to those used for training NMT models.

\begin{minipage}{\textwidth}
  \begin{minipage}[b]{0.45\textwidth}
    \centering
        \scalebox{0.59}{
    \begin{tabular}{c | c c |c c}
\toprule
Accuracy(\%) &\textbf{CATL} &\textbf{CATL-QTA} & \textbf{CATL}$^{W}$& \textbf{CATL-QTA}$^{W}$ \\
\midrule
Scene Recognition &93.18&93.45 &93.31&93.80\\
Sport Recognition &94.69 &95.45&94.96& 95.55\\
\rowcolor{blue!30} Color Attributes &54.66&56.08&57.59&60.16\\
Other Attributes & 48.52&50.30&52.25&54.36\\
\rowcolor{blue!30} Activity Recognition  &53.36&58.43&54.59 & 60.10\\
Positional Reasoning  &32.73&31.94&33.63 & 34.71\\
Sub. Object Recognition  &86.56&86.76& 86.52&86.98\\
Absurd &  95.03&100.00& 98.01& 100.00\\
Utility and Affordances &29.01&23.46&29.01 & 31.48\\
Object Presence &93.34&93.48&  94.13& 94.55\\
Counting  &50.08&49.93& 52.97&53.25\\
\rowcolor{blue!30} Sentiment Understanding  & 56.23&56.87& 62.62&64.38\\
\hline
Overall (Arithmetic MPT) &65.62&66.34&67.46 &69.11\\
Overall (Harmonic MPT) &55.95&54.60& 57.83& 60.08\\
\hline
Overall Accuracy &82.23&83.62& 83.92& 85.03\\
\bottomrule
\end{tabular}
}
      \captionof{table}{QTA in concatenation models on TDIUC dataset}
      \label{table:catcompare}
  \end{minipage}
  \hfill
  \begin{minipage}[b]{0.55\textwidth}
    \centering
    \includegraphics[width=5.5cm,height = 4cm]    {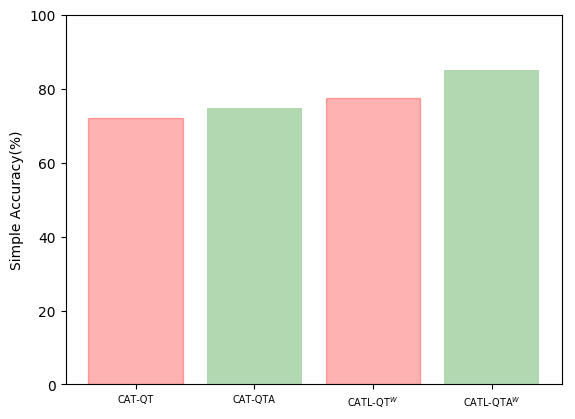}
      \captionof{figure}{Evaluation of different ways to utilize information from question type}
      \label{fig:information-attention}
    \end{minipage}
  \end{minipage}

\subsection{QTA in concatenation models}
\label{subsec:concat}
We use QTA in concatenation models to study the effect of QTA. The framework is in Figure~\ref{fig:qtype}. We compare the network using a weighted feature with the same network using an unweighted concatenated image feature in Table~\ref{table:catcompare}. As we can see, the model using the weighted feature has more power than the one using the unweighted feature. 9 out of 12 categories get improved results. ``Color" and ``Activity Recognition" get around 2$\%$ and 6$\%$ accuracy increases.

To ensure that the improvement is not because of the added question type information but the attention mechanism using question type, we show the comparison of QTA with QT in Figure~
\ref{fig:information-attention}. With same text feature and image feature and approximately same number of parameters in the network, QTA is 3-5$\%$ better than QT.

\begin{figure}[]
\centering
\includegraphics[width= 7cm, height= 6cm]{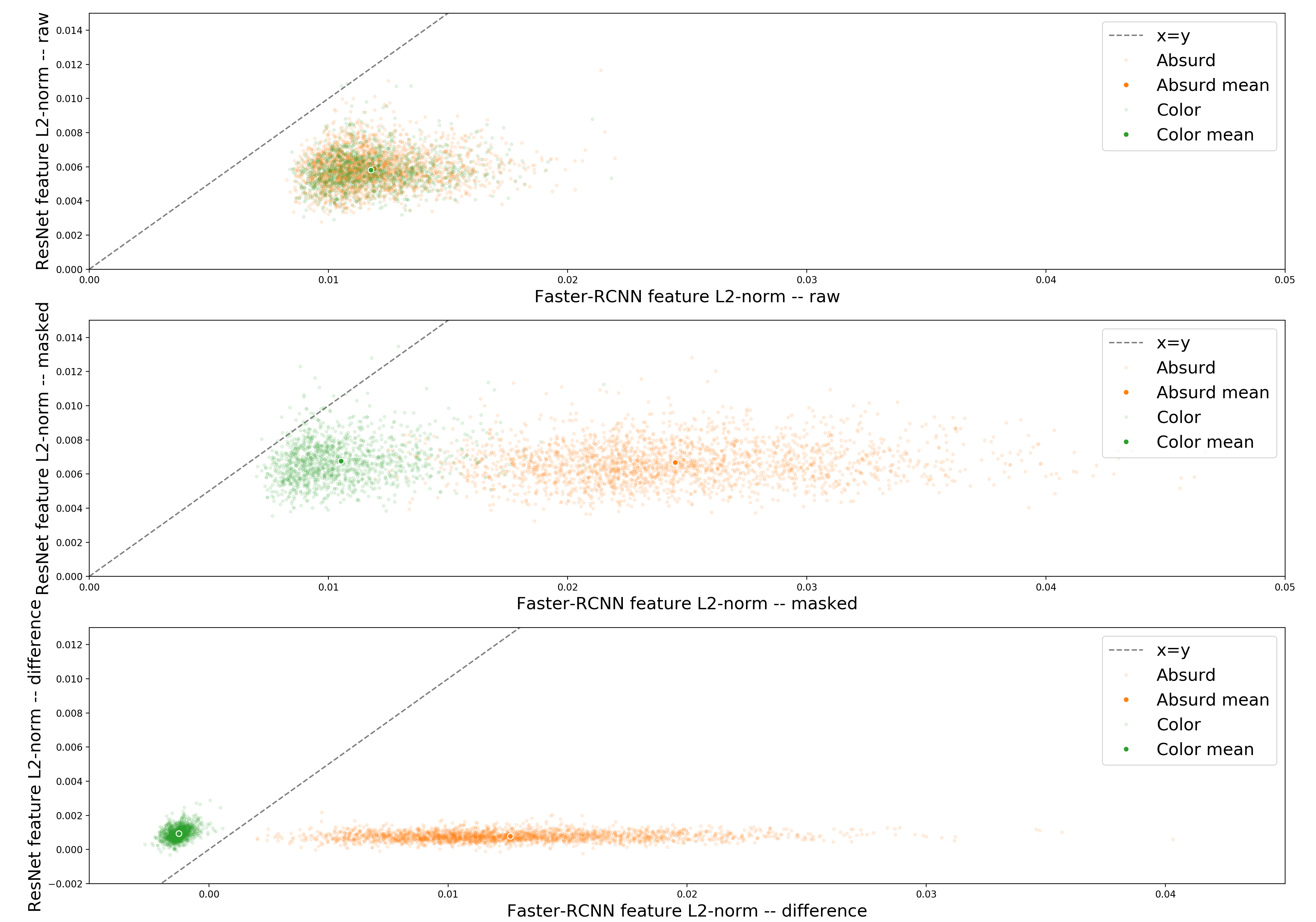}
\caption{Effects of weighting by QTA. Top: raw feature norms, Middle: feature norms weighted by QTA, Bottom: differences of norms after weighting vs before weighting. For color questions, the feature norms shift towards ResNet features, while for absurd questions they shift towards Faster-RCNN features.}
\label{fig:feature_norm}
\end{figure}

We show the effect of QTA on image feature norms in Figure~\ref{fig:feature_norm}. By weighing the image features by question type, we find that our model relies more on Faster R-CNN features for ``Absurd'' question samples while it relies more on ResNet features for ``Color'' questions.

The best setting we get in concatenation model is using a weighted image feature concatenated with the output of the LSTM and Word2Vec feature(CATL-QTA$^W$). It gets 5\% improvement compared to complicated deep network such as RAU and MCB-A in Table~\ref{Table:1}. 
\vspace*{-5pt}
\begin{table}[]
\caption{Results of QTA models on TDIUC dataset compared to state-of-art models}
\label{Table:1}
\begin{center}
\scalebox{0.8}{
\begin{tabular}{c | c c|| c c}
\toprule
Accuracy(\%) &  \textbf{CATL-QTA}$^{W}$ & \textbf{MCB-QTA}&\textbf{MCB-A}~\cite{TDIUC} & \textbf{RAU}~\cite{TDIUC}\\
\midrule
Scene Recognition & 93.80& 93.56&93.06&\textbf{93.96}\\
Sport Recognition & 95.55&\textbf{95.70} & 92.77&93.47\\
Color Attributes &60.16 & 59.82&\textbf{68.54} &66.86\\
Other Attributes & 54.36&54.06 & \textbf{56.72}&56.49\\
Activity Recognition & 60.10&\textbf{60.55} &52.35&51.60\\
Positional Reasoning &   34.71& 34.00 & \textbf{35.40}&35.26\\
Sub. Object Recognition & 86.98 & \textbf{87.00}& 85.54&86.11\\
Absurd & \textbf{100.00} & 100.00&84.82&96.08\\
Utility and Affordances &  31.48 &\textbf{37.04} & 35.09&31.58\\
Object Presence &\textbf{94.55} & 94.34&93.64&94.38\\
Counting & 53.25&\textbf{53.99} &51.01&48.43\\
Sentiment Understanding &64.38 & 65.65&\textbf{66.25}&60.09\\
\hline
Overall (Arithmetic MPT) & 69.11 &\textbf{69.69} &67.90&67.81\\
Overall (Harmonic MPT) & 60.08 & \textbf{61.56}&60.47&59.00\\
\hline
Overall Accuracy &\textbf{85.03} & 84.97&81.86&84.26\\
\bottomrule
\end{tabular}
}
\end{center}
\end{table}
\subsection{QTA in pooling models}
\label{subsec:pooling}
To show how to combine QTA with more complicated feature integration operator, we propose MCB-QTA structure. Even though MCB-QTA in Table~\ref{Table:1} doesn't win with simple accuracy, it shows great performance in many categories such as ``Object Recognition'' and ``Counting''. Accuracy in ``Utility and Affordances'' is improved by 6\% compared to our CATL-QTA model. It gets 8\% improvement in ``Activity recognition'' compared to state-of-art model MCB-A and also gets the best Arithmetic and Harmonic MPT value. 
\vspace*{-5pt}
\subsection{Multi-task analysis}
\label{subsec:multitask}
In this part, we will discuss how we use QTA when we have questions without specific question types. 
It is quite easy to predict the question type from the question itself. We use a 2-layer LSTM followed by a classifier and the test accuracy is 96\% after 9 epochs. The problem is whether we can predict the question type while keeping the same performance for VQA task or not.
As described in Figure~\ref{fig:multitask-qtype}, we use the predicted question type as input of the QTA network in a multi-task setting. We get 84.33\% test simple accuracy for VQA task as shown in Table~\ref{Table:5}. 
When we compare it to MCB-A or RAU in Table~\ref{Table:1}, though accuracy gets a little affected for most of the categories, we still get 2\% improvement in ``Sports Recognition'' and ``Counting''.

We fine-tune our model on VQA v1 using a pre-trained multi-task model that was trained on TDIUC. We use the question type predictor in the multi-task model as the input of QTA. Our model's performance is better than MCB in Table~\ref{Table:6} with an approximately same number of parameters in the network.

\begin{table}[]
\caption{Results of test-dev accuracy on VQA v1.
Models are trained on the VQA v1 train split and tested on test-dev}
\label{Table:6}
\begin{center}
\scalebox{0.8}{
\begin{tabular}{c | c}
\toprule
  \textbf{} &  \textbf{Accuracy(\%)}\\
\midrule
Element-wise Sum~\cite{MCB} & 56.50\\
Concatenation~\cite{MCB} &57.49\\
Concatenation + FC~\cite{MCB}& 58.40\\
Element-wise Product~\cite{MCB} &58.57\\
Element-wise Product + FC~\cite{MCB} &56.44\\
MCB(2048 $\times$ 2048 $\rightarrow$ 16K)~\cite{MCB}  &59.83\\
CATL-QTA-M + FC &\textbf{60.32}\\
\bottomrule
\end{tabular}
}
\end{center}
\end{table}
\vspace*{-5pt}
\subsection{Findings on TDIUC dataset}
\vspace*{-5pt}
To further analyze the effects of the question type prediction part in this multi-task framework, we list the confusion matrix for the question type prediction results in Table~\ref{table:cf1}. ``Color" and ``Absurd" question type predictions are most often bi-directionally confused. The reason for this is that among all absurd questions, more than 60\% are questions start with ``What color". To avoid this bias, we remove all absurd questions and run our multi-task model again. In this setting, our question type prediction did much better than before. Almost all categories get 99\% accuracy as shown in Table~\ref{table:cf2}.
We also compare our QTA models' performance without absurd questions in Table~\ref{Table:5}. In CATL-QTA network, removing absurd questions doesn't help much because in test we feed in the true question type labels. But it is useful when we consider the multi-task model. From third and fourth columns, we see that without absurd questions, we get improved performance among all categories. This is because we remove the absurd questions that may mislead the network to predict ``color" question type in the test.

\begin{table}
\centering
\caption{Confusion matrix for test question types prediction in CATL-QTA-M using TDIUC dataset. 1. Other Attributes 
2. Sentiment Understanding 
3. Sports Recognition
4. Position Reasoning 
5. Object Utilities/Affordances 
6. Activity Recognition 
7. Scene Classification 
8.  Color
9. Object Recognition 
10.Object Presence 
11.Counting 
12. Absurd}
\scalebox{0.8}{
\begin{tabular}{ c|cccccccccccc|c}
\toprule
 Target &
\multicolumn{12}{c}{Predicted} &  Acc(\%) \\
\midrule \midrule \\
  & 1 & 2 & 3 & 4 & 5 & 6 & 7 & 8 & 9 & 10 & 11 & 12 &95.66 \\
\cmidrule{2-13} 
 1 &\cellcolor{green!25}77.76& 0.00   & 0.89& 3.20 & 0.00     & 0.08& 0.42& 1.15&
        0.12& 0.00     & 0.00     & \cellcolor{red!25}16.38\\
 2  &0.80 & \cellcolor{green!25}60.51& 1.77& 8.83& 0.00    & 2.25& 2.57& 0.00     &
        1.44& 0.96& 0.16& \cellcolor{red!25}20.71\\
3 & 0.31& 0.00     & \cellcolor{green!25}73.08& 0.37& 0.00    & 0.17& 0.00     & 0.03&
        0.02& 0.00    & 0.01&\cellcolor{red!25} 26.01\\
4 &2.95& 0.02& 0.01& \cellcolor{green!25}89.52& 0.00     & 0.01& 0.02& 0.19&
        1.88& 0.03& 0.03& \cellcolor{red!25}5.35\\
5 &12.50 & 0.63& 3.12& \cellcolor{red!25}45.62&\cellcolor{green!25} 0.00     & 0.00     & 3.12& 0.00     &
        11.25& 0.00     & 0.00     & 23.75\\
6 &0.79& 0.00     & 14.56& 1.76& 0.00     & \cellcolor{green!25}13.18& 0.00     & 0.00     &
        2.21& 0.00     & 0.07& \cellcolor{red!25}67.43\\
7 &0.04& 0.00     & 0.04& \cellcolor{red!25}0.40 & 0.00     & 0.01&\cellcolor{green!25} 99.40 & 0.02&
        0.00     & 0.00     & 0.06& 0.03\\
8  &0.32& 0.00    & 0.18& 0.13& 0.00     & 0.00     & 0.00     & \cellcolor{green!25}86.10 &
        0.00    & 0.00     & 0.00     & \cellcolor{red!25}13.28\\
9 &0.01& 0.00     & 0.00     & 0.31& 0.00     & 0.00     & 0.00     & 0.00     &
    \cellcolor{green!25}    98.96& 0.01& 0.00     & \cellcolor{red!25}0.71\\
10 & 0.00    & 0.00     & 0.00     & 0.00     & 0.00     & 0.00     & 0.00     & 0.00     &
        0.00     &\cellcolor{green!25}100.00     & 0.00     & 0.00  \\
11 &  0.00     & 0.00     & 0.00     & 0.01& 0.00     & 0.00     & 0.02& 0.00     &
        0.02& \cellcolor{red!25}0.05& \cellcolor{green!25}99.90 & 0.00   \\
12 &0.35& 0.00     & 0.18& 0.41& 0.00    & 0.03& 0.00    & \cellcolor{red!25}3.18&
       0.40 & 0.00     & 0.00    &\cellcolor{green!25} 95.46\\
\bottomrule
\end{tabular}
}
\label{table:cf1}
\end{table}
\section{Conclusion}
We propose a question type-guided visual attention (QTA) network. We show empirically that with the question type information, models can balance between bottom-up and top-down visual features and achieve state-of-the-art performance. 
\begin{table}[h]
\centering
\caption{Confusion matrix for test question types prediction in CATL-QTA-M using TDIUC dataset without absurd questions. Numbers represent same categories as in Table~\ref{table:cf1}}
\scalebox{0.7}{
\begin{tabular}{ c|cccccccccccc|c}
\toprule
 Target &
\multicolumn{12}{c}{Predicted} &  Acc(\%) \\
\midrule \midrule \\
  & 1 & 2 & 3 & 4 & 5 & 6 & 7 & 8 & 9 & 10 & 11 & 12 &99.50\\
\cmidrule{2-13} 
 1 &\cellcolor{green!25}98.39& 0.00    & 0.07& 0.15& 0.00    & 0.13& 0.08& \cellcolor{red!25}0.63&
        0.55& 0.00    & 0.00    & N/A  \\
 2  &0.16&\cellcolor{green!25} 84.03& 3.67& 0.00    & 0.00    & 3.35& \cellcolor{red!25}5.59& 0.00   &
        0.48& 0.00   & 2.72& N/A \\
3 &0.00    & 0.08&\cellcolor{green!25} 97.31& 0.00    & 0.00    & \cellcolor{red!25}2.37& 0.01& 0.00    &
        0.10 & 0.02& 0.11& N/A  \\
4 &\cellcolor{red!25}1.01& 0.00    & 0.00    & \cellcolor{green!25}98.07& 0.00   & 0.01& 0.00    & 0.51&
        0.41& 0.00    & 0.00    &N/A\\
5 &8.64& 3.70 & 14.81& 0.00    & \cellcolor{green!25}0.00    & \cellcolor{red!25}59.26& 7.41& 1.23&
        4.94& 0.00    & 0.00    & N/A \\
6 &0.45& 0.15& \cellcolor{red!25}31.42& 0.00    & 0.00    & \cellcolor{green!25}67.39& 0.04& 0.04&
        0.45& 0.00   & 0.07& N/A\\
7 & 0.02& 0.03& 0.00    & 0.00    & 0.00    & 0.03& \cellcolor{green!25}99.86& 0.02&
        0.00    & 0.00    & \cellcolor{red!25}0.04& N/A\\
8  &   0.06& 0.00    & 0.00    & \cellcolor{red!25}0.13& 0.00    & 0.04& 0.07& \cellcolor{green!25}99.70 &
        0.00    & 0.00   & 0.00    & N/A \\
9 &0.06& 0.00    & \cellcolor{red!25}0.13& 0.01& 0.00    & 0.02& 0.00    & 0.00   &
      \cellcolor{green!25}  99.76& 0.01& 0.00    &N/A  \\
10 & 0.00    & 0.00    & 0.00    & 0.00    & 0.00    & 0.00    & 0.00    & 0.00    &
        0.00    &\cellcolor{green!25} 100.00    & 0.00    & N/A \\
11 &  0.00    & 0.00    & 0.01& 0.00    & 0.00    & 0.00    & 0.00    & 0.00    &
        0.00    & \cellcolor{red!25}0.01& \cellcolor{green!25}99.98& N/A\\
12 &N/A    & N/A     & N/A    & N/A    & N/A     & N/A     & N/A     & N/A     &
       N/A     & N/A     & N/A     & N/A\\
\bottomrule
\end{tabular}
}
\label{table:cf2}
\end{table}
\begin{table}[]
\caption{Results of test accuracy when question type is hidden with/without absurd questions in training. We compare them with similar QTA models. * denotes training and testing without absurd questions}
\label{Table:5}
\begin{center}
\scalebox{0.6}{
\begin{tabular}{c |c c c ||c c || c }
\toprule
  \textbf{} & \textbf{CATL-QTA}$^{W}$ &\textbf{CATL}$^{W*}$ & \textbf{CATL-QTA}$^{W*}$ & \textbf{CATL-QTA-M} & \textbf{CATL-QTA-M}$^{*}$ & \textbf{CAT1}$^{1*}$~\cite{TDIUC} \\
\midrule
Scene Recognition & 93.80&93.46 &93.62&93.74  &93.82 &72.75\\
Sport Recognition &95.55 &94.97 & 95.47&94.80&95.31&89.40\\
Color Attributes&60.16 &57.84&  58.63&57.62&59.73&50.52\\
Other Attributes &54.36 &53.90& 53.44&52.05 &56.17&51.47\\
Activity Recognition  &60.10 & 57.38& 59.43&53.13&58.61&48.55\\
Positional Reasoning  &34.71 & 33.98& 34.63&33.90&34.70&27.73\\
Sub. Object Recognition  &86.98& 86.62& 86.74&86.89&86.80&81.66\\
Absurd &100.00 &N/A & N/A&98.57&N/A& N/A\\
Utility and Affordances & 31.48 & 27.78 & 34.57&24.07&35.19&30.99\\
Object Presence  &94.55 & 93.87& 94.22& 94.57&94.60&69.50\\
Counting  &53.25 & 52.33& 52.20& 53.59&55.30&44.84\\
Sentiment Understanding&64.38 &64.06 & 65.81& 60.06&61.31&59.94\\
\hline
Overall (Arithmetic MPT) &69.11&65.11 & 66.25& 66.92&66.88&57.03\\
Overall (Harmonic MPT) &60.08 & 55.89& 58.51 & 55.77&58.82&50.30\\
\hline
Simple Accuracy & 85.03& 79.79 & 80.13& 84.33&80.95&63.30\\
\bottomrule
\end{tabular}
}
\end{center}
\end{table}
Our results show that QTA systematically improves the performance by more than 5\% across multiple question type categories such as ``Activity Recognition'', ``Utility'' and ``Counting'' on TDIUC dataset. 
We consider the case when we don't have question type for test and propose a multi-task model to overcome this limitation by adding question type prediction task in the VQA task. We get around 95\% accuracy for the question type prediction while keeping the VQA task accuracy almost same as before.\\
\textbf{Acknowledgements}
We thank Amazon AI for providing computing resources. Yang Shi is supported by Air Force Award FA9550-15-1-0221.
\clearpage

\bibliographystyle{splncs04}
\bibliography{egbib}
\end{document}